\documentclass[11pt]{article}

\usepackage[final]{acl}

\usepackage[table]{xcolor}   % colortbl 먼저 활성화
\usepackage{booktabs}
\usepackage{arydshln}        % booktabs 뒤
\usepackage{makecell}
\usepackage{adjustbox}

\newcommand{\eg}{\textit{e}.\textit{g}.}
\usepackage{siunitx}
\usepackage{amsmath}
\usepackage{amssymb}
\usepackage{multirow}
\usepackage{pifont}
\newcommand{\cmark}{\ding{51}}
\newcommand{\xmark}{\ding{55}}
\usepackage{arydshln}
\usepackage{tabularx}
\usepackage{times}
\usepackage{latexsym}

\usepackage[T1]{fontenc}
\usepackage[utf8]{inputenc}

\usepackage{microtype}

\usepackage{inconsolata}

\usepackage{graphicx}

\title{Diffusion Large Language Models for Visual Speech Recognition}

\author{Jeong Hun Yeo \quad Chae Won Kim \quad Hyeongseop Rha \quad Yong Man Ro$^*$
\\Integrated Vision Language Lab, KAIST, South Korea\\
\small{\texttt{\{sedne246, chaewonkim, ryool\textunderscore1832, ymro\}@kaist.ac.kr}}}

\begin{document}
\maketitle
\def\thefootnote{}\footnotetext{$^*$Corresponding Author.}

\begin{abstract}
Existing Visual Speech Recognition (VSR) systems commonly rely on left-to-right autoregressive decoding, which can force premature decisions on visually ambiguous tokens before sufficient context is available. We propose DLLM-VSR, to the best of our knowledge, the first Diffusion Large Language Model (DLLM)-based VSR framework, formulating transcription as iterative masked denoising with flexible-order decoding. With confidence-based unmasking, DLLM-VSR commits high-confidence positions early and uses the committed tokens as bidirectional context to refine ambiguous ones. To adapt DLLMs to VSR, we introduce a two-stage masked-denoising training strategy that separates visual-to-text content alignment from length modeling. We further observe a performance gap compared with an upper-bound setting where the ground-truth transcript length is provided at inference, allowing the model to focus on transcript content decoding. To reduce this gap, we develop length-guided candidate decoding, which uses video duration to construct plausible transcript-length hypotheses and reranks the decoded candidates using length plausibility and decoding confidence. The proposed method achieves a 19.4\% word error rate on LRS3, establishing state-of-the-art performance among methods using only LRS3 as labeled training data. The code is available at \url{https://bit.ly/DLLM-VSR}.
\end{abstract}

\section{Introduction}
Visual Speech Recognition (VSR)~\cite{afouras2018deep, ma2023auto}, also known as lip reading, aims to transcribe spoken utterances into text using only visual cues from a speaker’s mouth movements. As a visual-only speech interface, VSR can provide deaf and hard-of-hearing users with visual access to spoken language and enable silent communication in situations where audio input is impractical or undesirable. Despite these promising applications, VSR remains a challenging task due to the inherent phonetic ambiguity of lip movements. Although spoken English contains approximately 40 distinct phonemes, these phonemes collapse into only about 10--13 visually distinguishable units, commonly referred to as visemes~\cite{cappelletta2012phoneme, bear2016decoding}. As a result, multiple phonemes may belong to the same viseme group (\eg, /p/, /b/, and /m/), making them nearly indistinguishable from visual evidence alone.

\begin{figure}[t]
    \centering
    \includegraphics[width=\linewidth]{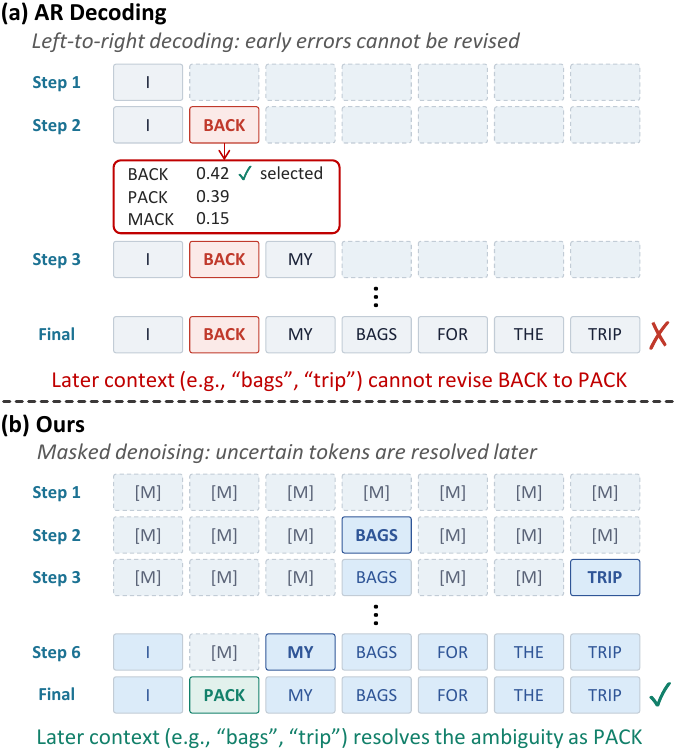}
    \caption{Conceptual comparison between autoregressive decoding and the proposed method.}
    \label{fig:1}
\end{figure}

To mitigate this ambiguity, recent VSR systems have improved through advances in both visual encoders and text decoders. On the encoder side, attention-based Transformer architectures have enabled long-range temporal modeling of lip movement sequences, while self-supervised pretraining has improved visual speech representations by leveraging large-scale unlabeled audio-visual data~\cite{shi2022learning, haliassos2023jointly}. On the decoder side, autoregressive decoders have become a common choice, generating transcripts token by token in a fixed left-to-right order~\cite{afouras2018deep, ma2023auto}. More recently, Large Language Model (LLM)-based decoders~\cite{grattafiori2024llama, Yang2024Qwen25TR} have further improved performance by injecting stronger linguistic priors into the decoding process~\cite{yeo2024visual, cappellazzo2025large}.

However, despite these gains, current LLM-based VSR systems remain constrained by the fixed left-to-right token generation order of autoregressive decoding. While suitable for sequential transcription, this rigid order can be suboptimal for VSR, where visual evidence is highly uneven across output positions. Due to viseme ambiguity and weak visual cues, some tokens may remain highly uncertain, whereas others can be predicted with relatively high confidence from visual evidence or contextual constraints. Consequently, strict left-to-right generation can force the model to commit to ambiguous early positions before easier, high-confidence positions become available as contextual anchors. This motivates a strategy that first commits high-confidence positions and progressively uses the committed tokens to disambiguate uncertain ones, as illustrated in Figure~\ref{fig:1}.

In this paper, we propose DLLM-VSR, the first Diffusion Large Language Model (DLLM)-based framework for VSR to the best of our knowledge. Unlike autoregressive decoders, DLLMs start from a fixed-length canvas initialized with mask tokens and iteratively denoise masked positions into text tokens, enabling flexible-order generation~\cite{nie2025large, ye2025dream}. To determine which positions to unmask and commit at each iteration, we adopt confidence-based unmasking, a common DLLM decoding strategy that prioritizes high-confidence positions~\cite{yu2025dimple}. This naturally matches VSR by operationalizing the high-confidence-first decoding strategy: reliable positions are unmasked early, and through bidirectional refinement, the committed tokens guide visually ambiguous positions.

To train the model, a standard DLLM formulation represents each target sequence on a fixed-length masked canvas with transcript tokens followed by an end-of-sequence (EOS) token and padding (PAD) tokens, enabling variable-length transcripts to be trained under a unified denoising objective. A direct instantiation for DLLM-VSR would therefore supervise transcript positions with text tokens and positions beyond EOS with PAD tokens. However, this can lead to padding-heavy supervision, as shorter transcripts leave many canvas positions assigned to padding prediction. Recent observations suggest that excessive padding loss can reduce sample efficiency and cause generation instability in DLLMs~\cite{xie2025dream}. Inspired by this observation, we propose a two-stage masked-denoising training strategy tailored to VSR, which separates content learning from length modeling: the first stage predicts only transcript tokens and the immediately following EOS token, while the second additionally predicts PAD tokens beyond EOS for length modeling.

After two-stage training, DLLM-VSR outperforms recent LLM-based VSR systems, showing the effectiveness of the proposed DLLM-based decoding framework for VSR. Furthermore, to investigate its performance upper bound, we evaluate the proposed method in an oracle setting where the ground-truth transcript length is given and used to fix EOS and PAD positions in advance. The substantial error reduction in this setting indicates that reducing uncertainty in EOS and PAD placement could further improve decoding. Motivated by this observation, we introduce length-guided candidate decoding, which leverages video duration to construct plausible transcript-length hypotheses, decodes candidates for each hypothesis, and selects the final transcript by reranking them using length plausibility and decoding confidence. Together, our training and decoding strategies achieve a 19.4\% word error rate (WER) on LRS3, establishing state-of-the-art performance among methods using only LRS3 as labeled training data.

\section{Related Work}
\subsection{Visual Speech Recognition}
VSR aims to transcribe silent lip movement videos into text. Modern VSR systems typically follow an encoder--decoder paradigm, where a visual encoder extracts visual speech representations from lip movements and a text decoder converts them into text tokens. On the encoder side, prior work has improved representations through visual front-ends~\cite{stafylakis2017combining}, Transformer- and Conformer-based temporal modeling~\cite{afouras2018deep, ma2021end}, audio-guided multimodal self-supervised learning~\cite{shi2022learning, haliassos2023jointly}, and data scaling with ASR-generated pseudo labels~\cite{yeo2024visual2, ma2023auto}. While these advances have strengthened encoder-side representations, this work focuses on the decoding stage, which remains crucial for resolving visual ambiguity.

Early VSR decoders were built on Connectionist Temporal Classification (CTC)~\cite{assael2016lipnet}, which offers efficient non-autoregressive decoding but relies on a conditional independence assumption that limits the modeling of dependencies among output tokens. Autoregressive decoders~\cite{afouras2018deep} address this limitation by incorporating language modeling into decoding, where each token is predicted conditioned on visual representations and previously generated tokens. Recent work has further leveraged abundant audio and audio-text data to strengthen decoder-side linguistic priors. Examples include transferring knowledge from pretrained ASR models~\cite{zhao2020hearing, ren2021learning}, leveraging paired audio-text data for decoder pretraining~\cite{kim2023lip, yeo2024akvsr}, and aligning visual representations with intermediate representations of strong speech models such as Whisper~\cite{prajwal2024speech}. More recently, LLM-based decoders~\cite{yeo2024visual, cappellazzo2025large} have been introduced to exploit even stronger linguistic priors for resolving visual ambiguity in VSR. However, most of these approaches still retain the standard left-to-right generation order, despite the highly ambiguous and uneven nature of visual speech evidence. In line with this decoder-centric direction, our work improves VSR not by strengthening linguistic priors alone, but by revisiting the token generation order itself.

\begin{figure*}[t]
    \centering
    \includegraphics[width=0.99\textwidth]{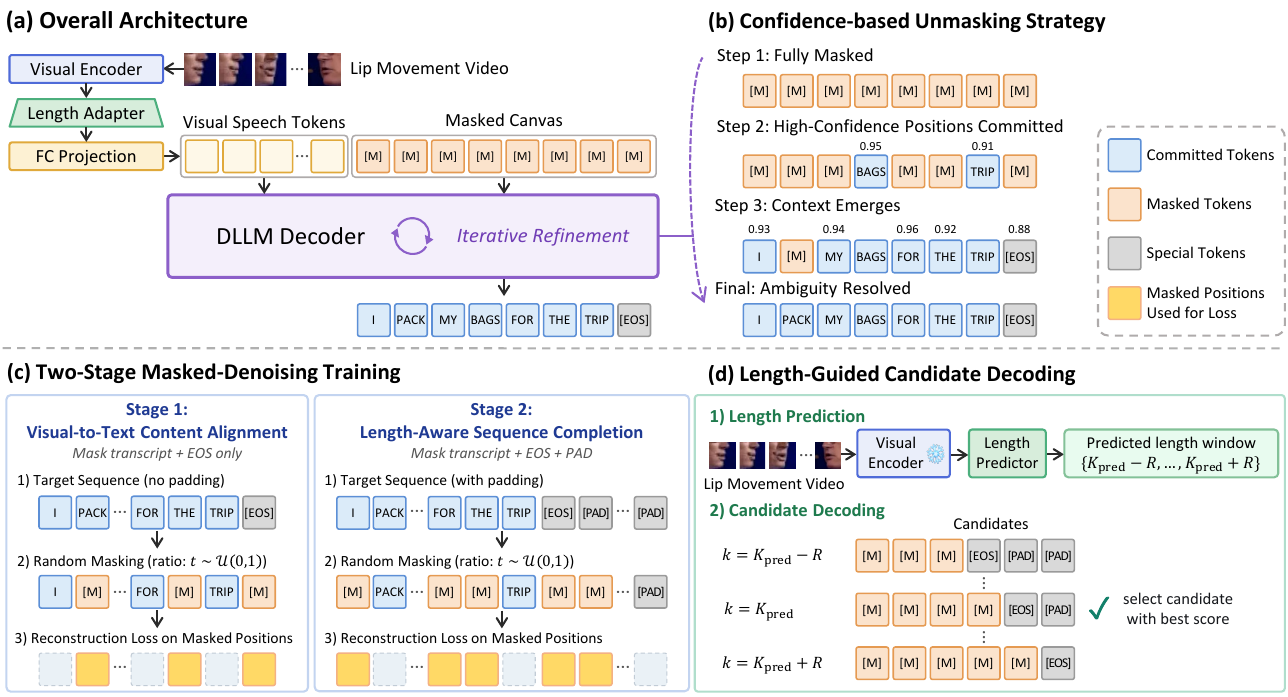}
    \caption{
    Overview of DLLM-VSR.
    (a) Overall architecture with a frozen visual encoder, a length adapter, FC projection layers, and a LoRA-adapted DLLM decoder.
    (b) Confidence-based unmasking, where high-confidence positions are committed first and the committed tokens are used as bidirectional textual context.
    (c) Two-stage masked-denoising training for visual-to-text content alignment and length-aware sequence completion. Yellow boxes indicate masked positions used for the reconstruction loss.
    (d) Length-guided candidate decoding with multiple length hypotheses and joint candidate reranking. $R$ denotes the candidate-length search radius, i.e., candidate lengths are decoded within $K_{\mathrm{pred}} \pm R$.
    }
    \label{fig:2}
\end{figure*}

\subsection{Diffusion Large Language Models}
DLLMs have recently attracted increasing attention for enabling parallel decoding with bidirectional attention, offering a promising direction for improving LLM inference efficiency. Early scaling efforts such as LLaDA~\cite{nie2025large} demonstrated that diffusion-based language modeling can be trained from scratch at the 8B scale and achieve performance competitive with strong autoregressive LLMs. Another line of work adapts pretrained autoregressive LLMs into diffusion language models, from earlier efforts such as DiffuLLaMA and DiffuGPT~\cite{gong2025scaling} to recent models such as Dream~\cite{ye2025dream}, which builds on Qwen2.5~\cite{Yang2024Qwen25TR} and shows strong performance on general language, mathematical reasoning, and code generation tasks. More recently, multimodal DLLMs such as LLaDA-V~\cite{you2025llada}, LaViDa~\cite{li2025lavida}, and Dimple~\cite{yu2025dimple} have extended diffusion-style decoding to visual inputs through visual instruction tuning or adaptation from pretrained vision-language models. These advances suggest that DLLMs are evolving from text-only generators into general-purpose decoders for multimodal conditional generation.

Despite this progress, DLLMs face a fundamental challenge in variable-length generation due to their fixed-length canvas. Unlike autoregressive models, which terminate by emitting an EOS token, DLLMs must commit to a canvas size before denoising. A too-short canvas may truncate the output, whereas an overly long canvas wastes computation on redundant EOS or padding positions and can degrade generation quality. Existing approaches address this issue through three main directions: training-based length control~\cite{yang2025diffusion,wu2026dreamon}, such as DreamOn’s [expand]/[delete] operations; training-free canvas adaptation based on inference-time signals~\cite{li2026beyond, cheng2026improving}, such as DAEDAL’s EOS-confidence criterion; and semi-autoregressive block-wise decoding, as in Block Diffusion~\cite{arriola2025block}. Our approach follows the training-free direction, but differs in how the target length is estimated. Instead of relying solely on model-internal signals to search for an adequate canvas size, we exploit a task-specific property of VSR: transcript length is strongly correlated with input video duration.
\section{Method}
Our model is illustrated in Figure~\ref{fig:2}. Given a lip movement video and a masked transcript canvas, the model generates a transcript by iteratively denoising masked positions, where high-confidence positions are committed and their tokens serve as textual context for resolving visually ambiguous ones. We first describe the visual-conditioned DLLM architecture, then introduce our two-stage masked-denoising training strategy and length-guided candidate decoding.

\subsection{Architecture}
Following recent LLM-based VSR systems~\cite{yeo2024visual, cappellazzo2025large}, we retain the visual encoder and projection interface but replace the left-to-right autoregressive decoder with a DLLM decoder. Given a lip movement video $V = \{f_1, \dots, f_N\}$ consisting of $N$ frames, our goal is to generate the transcript $x_0 = \{x_0^1, \dots, x_0^K\}$ of length $K$. The architecture consists of a pretrained visual encoder~\cite{shi2022learning, haliassos2026pay}, a length adapter, two projection layers that map visual features into the language embedding space, and a DLLM decoder conditioned on the resulting visual speech tokens $v$.

A DLLM models the transcript on a fixed-length canvas of $T$ token positions, where $T$ is chosen to accommodate the longest training transcript including the EOS token. Starting from a fully masked canvas, the DLLM iteratively predicts token distributions for all masked positions conditioned on the visual speech tokens $v$ and the currently unmasked tokens. Following confidence-based unmasking~\cite{yu2025dimple}, we commit positions whose confidence exceeds a fixed threshold, or the most confident position if none exceeds the threshold. Once a position is committed, its predicted token is kept fixed and used as bidirectional context for the remaining masked positions.

To train the DLLM decoder, we follow the masked diffusion formulation~\cite{nie2025large}. Given a target sequence $x_0$ on the canvas, we sample a masking ratio $t \sim \mathcal{U}(0,1)$ and replace selected tokens with the mask token $\mathrm{M}$ to obtain $x_t$. Let $\mathcal{M} = \{i : x_t^{i} = \mathrm{M}\}$ denote the masked positions. Conditioned on the visual speech tokens $v$ and the unmasked tokens in $x_t$, the model reconstructs the original tokens at masked positions using the following loss:
\begin{equation}
\setlength{\abovedisplayskip}{3pt}
\setlength{\belowdisplayskip}{3pt}
\label{eq:loss}
\mathcal{L} =
-\mathbb{E}_{t, v, x_0, x_t}
\left[
\frac{1}{t}
\sum_{i \in \mathcal{M}}
\log p_{\theta}(x_0^{i} \mid v, x_t)
\right],
\end{equation}
where $\theta$ denotes the trainable parameters. Our two-stage strategy below uses the same objective but differs in the target canvas and the set of positions included in $\mathcal{M}$.

\subsection{Two-Stage Masked-Denoising Training}
A straightforward way to train the DLLM decoder is to apply the diffusion objective directly over the full canvas containing transcript, EOS, and PAD tokens. However, when the canvas is much longer than the transcript, PAD tokens occupy a large fraction of supervised positions and can dominate the loss. We therefore decompose training into two stages: the first learns content prediction using only transcript and EOS positions, while the second extends denoising to the full canvas to learn padding completion beyond EOS. Since the EOS token is attached directly after the transcript, the first stage still exposes the model to transcript termination without requiring it to model a long non-content suffix.

\subsubsection{Stage 1: Visual-to-Text Content Alignment}
In the first stage, we exclude padding positions and train on the transcript followed by an EOS token. Given $x_0 = \{x_0^1, \dots, x_0^K, \mathrm{EOS}\}$, we sample a masking ratio $t \sim \mathcal{U}(0,1)$ and mask its tokens to obtain $x_t$, so that $\mathcal{M}$ spans only the transcript tokens and the following EOS token. Training with Eq.~\ref{eq:loss} over $\mathcal{M}$ encourages visual-to-text content alignment and transcript termination without being dominated by padding prediction.

\subsubsection{Stage 2: Length-Aware Sequence Completion}
Building on Stage~1, the second stage extends each sequence to the canvas length $T$ by filling positions after the EOS token with PAD tokens. We then apply masked reconstruction over the entire canvas, so that $\mathcal{M}$ spans transcript, EOS, and PAD positions. Training with Eq.~\ref{eq:loss} over this full-canvas $\mathcal{M}$ teaches the model to preserve transcript and EOS prediction while completing the remaining canvas with PAD tokens, enabling variable-length generation within a fixed-length canvas.

\subsection{Length-Guided Candidate Decoding}
A DLLM generates a transcript by denoising a fixed-length canvas of $T$ positions, where the transcript occupies a prefix and the remaining positions are filled with an EOS token and PAD tokens. Thus, the transcript length $K$ determines the partition between content and non-content positions. Inferring $K$ implicitly during denoising can be suboptimal for VSR: overestimated lengths introduce spurious content positions, while underestimated lengths truncate the transcript. We therefore propose length-guided candidate decoding, which predicts plausible transcript lengths from the input video and decodes candidates under multiple length hypotheses.

\noindent\textbf{Length prediction.}
We attach a lightweight length predictor on top of the frozen visual encoder to estimate the transcript length $K$ from the visual feature sequence. Since the visual features are produced at a fixed temporal rate, their sequence length reflects the input video duration and provides a useful cue for transcript length. The predictor pools the visual features with a learnable query token and classifies over candidate lengths, yielding $P(K \mid v)$. It is trained independently using cross-entropy loss against the ground-truth transcript length.

\noindent\textbf{Candidate decoding under length hypotheses.}
Rather than relying on a single predicted length, we decode a candidate pool around the top-1 prediction, $\mathcal{K} = \{K_{\text{pred}} - R, \dots, K_{\text{pred}} + R\}$, where $K_{\text{pred}} = \arg\max_k P(k \mid v)$ and $R$ denotes the candidate-length search radius. For each $k \in \mathcal{K}$, we initialize the first $k$ positions as mask tokens, pin an EOS token at position $k+1$, and fill the remaining positions with PAD tokens. The pinned EOS and PAD tokens are kept fixed throughout decoding, and confidence-based unmasking is applied only to the first $k$ transcript positions. For efficiency, we batch the candidates corresponding to all $k \in \mathcal{K}$ and perform denoising in parallel. Decoding each length hypothesis yields a candidate transcript, together with the token confidence values and the number of denoising iterations required for completion, which are used for joint reranking.

\noindent\textbf{Joint reranking.}
We select the final transcript by reranking candidates with the following joint score:
\begin{equation}
\setlength{\abovedisplayskip}{3pt}
\setlength{\belowdisplayskip}{3pt}
\label{eq:rerank}
s(k) =
\sum_{i=1}^{k} \log c_i
+ \lambda \log p_k
- \beta n_k,
\end{equation}
where $c_i$ is the confidence of the committed token at transcript position $i$, $p_k = P(k \mid v)$ is the probability assigned to length $k$ by the length predictor, $n_k$ is the number of denoising iterations required to complete the candidate of length $k$, and $\lambda$ and $\beta$ are balancing weights. We refer to the three terms as the \emph{sequence confidence score}, \emph{length-prediction score}, and \emph{iteration penalty}, respectively. The sequence confidence score aggregates the log-confidence of tokens at the time they are committed and is used as a sequence-level confidence measure for candidate comparison. The length-prediction score favors candidate lengths assigned high probability by the video-conditioned length predictor. The iteration penalty penalizes candidates according to the number of denoising iterations required for completion. The final transcript is obtained from the highest-scoring candidate:
\begin{equation}
\setlength{\abovedisplayskip}{3pt}
\setlength{\belowdisplayskip}{3pt}
\label{eq:kstar}
k^{*} = \arg\max_{k \in \mathcal{K}} s(k).
\end{equation}
\begin{table*}[t]
\centering
\small
\setlength{\tabcolsep}{3.4pt}
\renewcommand{\arraystretch}{1.18}
\begin{adjustbox}{max width=0.8\linewidth}
\begin{tabular}{@{}llllrrr@{}}
\toprule
\textbf{Method} 
& \textbf{Pub.}
& \textbf{Encoder} 
& \textbf{Decoder} 
& \makecell{\textbf{Encoder}\\\textbf{Pretraining}\\\textbf{Data (h)}}
& \makecell{\textbf{Labeled}\\\textbf{Data (h)}} 
& \makecell{\textbf{WER}\\\textbf{(\%)$\downarrow$}} \\
\midrule
\rowcolor{gray!12}
\multicolumn{7}{@{}c@{}}{\textit{Fully Supervised}} \\
\citet{afouras2018deep}        & TPAMI'18       & Transformer       & Transformer & --    & 1,519   & 58.9 \\
\citet{makino2019recurrent}    & ASRU'19        & BiLSTM       & RNN-T       & --    & 31,000  & 33.6 \\
\citet{prajwal2022sub}         & CVPR'22        & Transformer       & Transformer & --    & 2,676   & 30.7 \\
\citet{ma2023auto}             & ICASSP'23      & Conformer & Transformer & --    & 3,448   & 19.1 \\
\citet{serdyuk2022transformer} & Interspeech'22 & ViT3D     & RNN-T       & --    & 90,000  & 17.0 \\
\citet{chang2024conformer}     & ICASSP'24      & Conformer & RNN-T       & --    & 100,000 & 12.8 \\
\midrule
\rowcolor{gray!12}
\multicolumn{7}{@{}c@{}}{\textit{Self-Supervised}} \\
\citet{shi2022learning}        & ICLR'22        & AV-HuBERT & Transformer & 1,759 & 433 & 28.6 \\
\citet{zhu2023vatlm}           & TMM'24         & VATLM & Transformer & 1,759 & 433 & 28.4 \\
\citet{haliassos2023jointly}   & ICLR'23        & RAVEn     & Transformer & 1,759 & 433 & 27.8 \\
\citet{haliassos2026pay}       & ICLR'26        & USR 2.0   & Transformer & 3,305 & 433 & $^{\ddagger}$ 25.8 \\
\midrule
\rowcolor{gray!12}
\multicolumn{7}{@{}c@{}}{\textit{LLM-Augmented}} \\
\citet{yeo2024visual}          & EMNLP'24       & AV-HuBERT & Llama2-7B    & 1,759 & 433 & 25.4 \\
\citet{cappellazzo2025large}   & ICASSP'25      & AV-HuBERT & Llama3-8B    & 1,759 & 433 & 25.3 \\
\citet{cappellazzo2026omni}    & ICASSP'26      & AV-HuBERT & Qwen2.5-7B   & 1,759 & 433 & 24.9 \\
\citet{Yuan_2025_ICCV}         & ICCV'25        & AV-HuBERT & Llama3-8B    & 1,759 & 433 & $^{\dagger}$ 22.0 \\
\midrule
\rowcolor{gray!12}
\multicolumn{7}{@{}c@{}}{\textit{DLLM-Augmented}} \\
Ours                  & --             & AV-HuBERT & Dream-7B     & 1,759 & 433 & 21.9 \\
Ours                  & --             & USR 2.0   & Dream-7B     & 3,305 & 433 & 19.4 \\
\bottomrule
\end{tabular}
\end{adjustbox}
\caption{Comparison with state-of-the-art VSR methods on LRS3. $^{\dagger}$ indicates the use of additional video-level metadata, such as title, description, and speaker name. $^{\ddagger}$ indicates our reproduction.}
\label{table:1}
\end{table*}

\section{Experimental Setup}
\subsection{Datasets}
We evaluate on three sentence-level English VSR benchmarks: LRS3~\cite{afouras2018lrs3}, LRS2~\cite{afouras2018deep}, and WildVSR~\cite{djilali2024vsr}. LRS3 is a widely used TED/TEDx-based VSR benchmark and serves as our primary dataset. Our LRS3 and WildVSR evaluations use models trained for VSR using only the 433-hour LRS3 training set as labeled data. LRS2, collected from BBC television programs, is used as a complementary benchmark, with all systems evaluated on LRS2 trained on its 223-hour training set. WildVSR is used for out-of-domain evaluation.

\subsection{Implementation Details}
We use two self-supervised pretrained visual encoders, USR~2.0 Huge~\cite{haliassos2026pay} and AV-HuBERT Large~\cite{shi2022learning}, with Dream-7B~\cite{ye2025dream} as the DLLM decoder. Visual features are downsampled from 25 to 12.5~fps by a 1D convolutional length adapter and mapped to Dream-7B's 3584-dimensional hidden space by a 2-layer projector. LoRA is applied to all linear layers with rank $r=16$ and $\alpha=32$. Length-guided candidate decoding uses a lightweight 2-layer Transformer length predictor with roughly 4M trainable parameters. Detailed architecture-, encoder-, and dataset-specific optimization settings are provided in Appendix~\ref{appendix:impl}. We report word error rate (WER) to measure recognition accuracy and real-time factor (RTF) to evaluate decoding efficiency, where RTF is defined as decoding time divided by video duration. RTF is measured on 50 test videos with batch size 1 on a single GPU, following 5 warm-up samples that are excluded from measurement.

\subsection{Inference and Decoding Strategies}
To characterize the role of transcript-length information at inference, we consider three decoding strategies. For \textit{oracle-length decoding}, the ground-truth sequence length is given, providing an upper bound with respect to transcript-length uncertainty. Decoding is performed on a length-matched canvas that exactly fits the transcript and its terminating EOS token. For \textit{implicit-length decoding}, the transcript length is unknown, and decoding starts from a fully masked canvas of length $T=32$. The model jointly predicts the transcript content and length; once an EOS token is committed, the positions after it are filled with PAD tokens and excluded from the output, while decoding continues until all remaining masked positions are committed. For \textit{length-guided candidate decoding}, we decode a pool of length hypotheses around the predicted length. For each hypothesis, transcript positions are initialized with mask tokens and the corresponding EOS position is fixed; for parallel decoding, shorter candidates are padded with PAD tokens to the length of the longest candidate in the batch. The final transcript is selected among the candidates within a search radius $R$ of $K_{\mathrm{pred}}$ using the joint score in Eq.~\ref{eq:rerank}. Reranking weights $\lambda$ and $\beta$ are selected on the validation set by grid search over $[0,1]\times[0,1]$ with a step size of 0.1. The selected values are $(0.9,0.7)$ for AV-HuBERT/LRS3, $(0.9,0.6)$ for USR~2.0/LRS3, and $(1.0,0.9)$ for USR~2.0/LRS2. For WildVSR, we use a canvas length of $T=96$ to accommodate longer utterances, while directly using the LRS3-trained model and length predictor with all other LRS3-tuned decoding and reranking settings, without adaptation. Unless otherwise specified, main results are reported with $R=3$ (see Sec.~\ref{sec:radius}). Across all decoding strategies, positions with confidence above 0.9 are committed; if none exceeds the threshold, the most confident position is committed.
\section{Experimental Results}
\begin{table}[t]
\centering
\small
\setlength{\tabcolsep}{4pt}
\renewcommand{\arraystretch}{1.2}
\begin{adjustbox}{max width=\linewidth}
\begin{tabular}{@{}llcccc@{}}
\toprule
\multirow{2}{*}{\textbf{Method}} &
\multirow{2}{*}{\textbf{Decoding}} &
\multirow{2}{*}{\makecell{\textbf{Bi-dir.}\\\textbf{Attn.}}} &
\multirow{2}{*}{\makecell{\textbf{Block}\\\textbf{Size}}} &
\multicolumn{2}{c}{\textbf{WER (\%)}$\downarrow$} \\
\cmidrule(lr){5-6}
& & & & \textbf{AV-HuBERT} & \textbf{USR~2.0} \\
\midrule
\citet{cappellazzo2026omni} & AR & \xmark & 1 & 24.9 & -- \\
\midrule
\rowcolor{gray!12}
\multicolumn{6}{@{}c@{}}{\textit{Implicit-length decoding without length guidance}} \\
\multirow{6}{*}{Ours}
   & AR       & \cmark & 1  & 24.3 & 22.2 \\
\cdashline{2-6}
   & \multirow{4}{*}{Block} & \cmark & 2  & 23.8 & 21.6 \\
   &                          & \cmark & 4  & 23.5 & 21.1 \\
   &                          & \cmark & 8  & 23.3 & 20.7 \\
   &                          & \cmark & 16 & 23.2 & 20.6 \\
\cdashline{2-6}
   & Parallel & \cmark & 32 & 23.1 & 20.5 \\
\bottomrule
\end{tabular}
\end{adjustbox}
\caption{Effect of token commitment order and bidirectional attention on LRS3. Block size denotes the number of mask tokens in each block; a block size of 32 corresponds to full-parallel decoding over the entire masked canvas.}
\label{table:2}
\end{table}

\subsection{Comparison with the State-of-the-Art Methods}
Table~\ref{table:1} compares the proposed method with state-of-the-art VSR methods on LRS3 across fully supervised, self-supervised, and LLM-based settings. Since our method builds on a self-supervised pretrained visual encoder, we first compare it with self-supervised encoder-based VSR models to assess the effect of replacing the autoregressive decoder with our DLLM-based framework. AV-HuBERT with a Transformer decoder reaches 28.6\% WER when trained for VSR using only LRS3 as labeled data. Coupling the same AV-HuBERT encoder with an LLM reduces WER; in particular, \citet{cappellazzo2026omni} combine AV-HuBERT with a Qwen2.5-7B LLM decoder and achieve 24.9\% WER. Our AV-HuBERT-based variant is directly comparable to this baseline, as Dream-7B is initialized from Qwen2.5-7B and further trained with masked denoising. Replacing the LLM decoder with a DLLM decoder further reduces WER from 24.9\% to 21.9\%.

We further evaluate our method with the USR~2.0 encoder. Since no prior work reports USR 2.0-based VSR using only LRS3 as labeled training data, we reproduce it with the encoder frozen, obtaining 25.8\% WER. Integrating this encoder into our framework reduces WER from 25.8\% to 19.4\%. WER reductions are observed with both visual encoders, indicating that the effectiveness of the proposed framework is consistent across encoder choices. The resulting 19.4\% WER approaches the 19.1\% WER of Auto-AVSR, which is trained on 3,448 hours of labeled data, while our model uses 433 hours of labeled LRS3 data.

\begin{table}[t]
\centering
\small
\setlength{\tabcolsep}{4pt}
\renewcommand{\arraystretch}{1.2}
\begin{adjustbox}{max width=\linewidth}
\begin{tabular}{@{}lcc@{}}
\toprule
\multirow{2}{*}{\textbf{Configuration}}
& \multicolumn{2}{c}{\textbf{WER (\%)}$\downarrow$} \\
\cmidrule(l){2-3}
& \textbf{AV-HuBERT} & \textbf{USR~2.0} \\
\midrule

\rowcolor{gray!12}
\multicolumn{3}{@{}c@{}}{\textit{Stage-1-only training}} \\
Oracle-length decoding   & 21.9  & 17.8 \\
Implicit-length decoding & 188.0 & 275.0 \\

\addlinespace[2pt]
\rowcolor{gray!12}
\multicolumn{3}{@{}c@{}}{\textit{After Stage~2 training}} \\
Oracle-length decoding (upper bound) & 20.2 & 17.7 \\
\hdashline
Implicit-length decoding (no length guidance) & 23.1 & 20.5 \\

\addlinespace[2pt]
\rowcolor{gray!12}
\multicolumn{3}{@{}c@{}}{\textit{Length-guided candidate decoding: reranking terms}} \\
Sequence confidence ($\sum \log c_i$)
+ length prediction ($\lambda \log p_k$)
& 22.6 & 20.2 \\
\quad + Iteration penalty ($-\beta n_k$)
& 21.9 & 19.4 \\

\bottomrule
\end{tabular}
\end{adjustbox}
\caption{Ablation of Stage~2 training and reranking terms in length-guided candidate decoding on LRS3.}
\vspace{-2pt}
\label{table:3}
\end{table}
\subsection{Effect of Token Commitment Order}
Table~\ref{table:2} analyzes the effect of token commitment order by first isolating bidirectional attention and then varying the block size. In block-based decoding, token positions within each block are committed based on confidence, and decoding proceeds to the next block only after the current block is fully decoded. Larger blocks therefore allow a wider range of positions to be committed in a flexible order. Compared with the LLM-based VSR method~\cite{cappellazzo2026omni}, our DLLM with autoregressive decoding reduces WER from 24.9\% to 24.3\% under the same left-to-right commitment order, showing the benefit of bidirectional attention. Performance consistently improves as the block size increases from 2 to 32, reducing WER from 23.8\% to 23.1\% with AV-HuBERT and from 21.6\% to 20.5\% with USR~2.0; full-parallel decoding performs best.

\subsection{Ablation Study}
Table~\ref{table:3} analyzes the effect of Stage~2 training and the contribution of the reranking terms used in length-guided candidate decoding on LRS3. We first examine the role of Stage~2 by evaluating a Stage-1-only model. Under oracle-length decoding, the Stage-1-only model achieves 21.9\% WER with AV-HuBERT and 17.8\% with USR~2.0. Under implicit-length decoding, its WER is 188.0\% and 275.0\%, respectively. After Stage~2 training, WER under the same implicit-length decoding setting is 23.1\% with AV-HuBERT and 20.5\% with USR~2.0. Oracle-length decoding achieves 20.2\% and 17.7\% WER, respectively, compared with 23.1\% and 20.5\% under implicit-length decoding. This gap represents the performance loss when transcript length must be inferred rather than provided. We then evaluate how the reranking terms in length-guided candidate decoding mitigate this gap. Using the sequence confidence and length-prediction scores in Eq.~\ref{eq:rerank} reduces WER from 23.1\% to 22.6\% with AV-HuBERT and from 20.5\% to 20.2\% with USR~2.0. Adding the iteration penalty further reduces WER to 21.9\% and 19.4\%, respectively.

\begin{table}[t]
\centering
\small
\setlength{\tabcolsep}{4pt}
\renewcommand{\arraystretch}{1.2}
\begin{adjustbox}{max width=\linewidth}
\begin{tabular}{@{}lcc@{}}
\toprule
\textbf{Model / Decoding}
& \textbf{WER (\%)}$\downarrow$
& \textbf{RTF}$\downarrow$ \\
\midrule

\rowcolor{gray!12}
\multicolumn{3}{@{}c@{}}{\textit{Autoregressive baselines}} \\
Transformer decoder (greedy)   & 24.7 & 0.09 \\
Qwen2.5-7B AR (greedy)         & 21.9 & 0.27 \\
Qwen2.5-7B AR (beam=5)         & 20.5 & 0.34 \\

\addlinespace[2pt]
\rowcolor{gray!12}
\multicolumn{3}{@{}c@{}}{\textit{Proposed method}} \\
Oracle-length decoding (upper bound) & 14.7 & 0.12 \\
\hdashline
Implicit-length decoding (no length guidance) & 17.7 & 0.14 \\
Length-guided candidate decoding             & 16.8 & 0.86 \\

\bottomrule
\end{tabular}
\end{adjustbox}
\caption{WER and decoding efficiency on LRS2. All models are trained and evaluated on LRS2 using the USR~2.0 visual encoder.}
\label{table:4}
\end{table}

\subsection{Effectiveness Across Datasets}
To verify that the effectiveness of the proposed method is not specific to LRS3, we further evaluate it on LRS2. The Transformer decoder obtains 24.7\% WER with an RTF of 0.09. Replacing it with a Qwen2.5-7B LLM decoder reduces WER to 21.9\% with greedy decoding and 20.5\% with beam search, with RTFs of 0.27 and 0.34, respectively. Using the proposed method, oracle-length decoding achieves 14.7\% WER with an RTF of 0.12, while implicit-length decoding obtains 17.7\% WER with an RTF of 0.14. Length-guided candidate decoding further reduces WER to 16.8\% with an RTF of 0.86. Consistent with the results on LRS3, the proposed method achieves lower WER than the autoregressive baselines on LRS2.

\subsection{Out-of-Domain Generalization to WildVSR}
To examine out-of-domain generalization, we compare the proposed method with AV-HuBERT and VSP-LLM on WildVSR. Compared with AV-HuBERT, VSP-LLM reduces WER from 28.6\% to 25.4\% on LRS3, while the WER on WildVSR changes only from 51.7\% to 51.6\%. Using the same AV-HuBERT encoder, the proposed method achieves 21.9\% WER on LRS3 and 46.1\% on WildVSR. Unlike the LLM-based baseline, the improvement observed on LRS3 is also observed on WildVSR. With the USR~2.0 encoder, WER is further reduced to 19.4\% on LRS3 and 39.9\% on WildVSR. Overall, the proposed method achieves lower WER on both LRS3 and WildVSR across the two visual encoders.

\begin{table}[t]
\centering
\small
\setlength{\tabcolsep}{3pt}
\renewcommand{\arraystretch}{1.2}
\begin{adjustbox}{max width=\linewidth}
\begin{tabular}{@{}lllcc@{}}
\toprule
\multirow{2}{*}{\textbf{Method}}
& \multirow{2}{*}{\textbf{Encoder}}
& \multirow{2}{*}{\textbf{Decoder}}
& \multicolumn{2}{c}{\textbf{WER (\%)}$\downarrow$} \\
\cmidrule(l){4-5}
& & & \textbf{LRS3} & \textbf{WildVSR} \\
\midrule

AV-HuBERT~\cite{shi2022learning}
& AV-HuBERT
& Transformer
& 28.6 & 51.7 \\

VSP-LLM~\cite{yeo2024visual}
& AV-HuBERT
& Llama2-7B
& 25.4 & 51.6 \\

\midrule

\multirow{2}{*}{Ours}
& AV-HuBERT
& \multirow{2}{*}{Dream-7B}
& 21.9 & 46.1 \\
& USR 2.0
&
& 19.4 & 39.9 \\

\bottomrule
\end{tabular}
\end{adjustbox}
\caption{Out-of-domain generalization from LRS3 to WildVSR. The proposed method uses the LRS3-trained model and length predictor with $T=96$ and all other LRS3-tuned decoding and reranking hyperparameters unchanged. WildVSR results for \citet{shi2022learning} and \citet{yeo2024visual} are reported by \citet{jain2026hype}.}
\label{table:5}
\vspace{-2pt}
\end{table}
\begin{table}[t]
\centering
\small
\setlength{\tabcolsep}{5pt}
\renewcommand{\arraystretch}{1.15}
\begin{adjustbox}{max width=\linewidth}
\begin{tabular}{@{}lcccc@{}}
\toprule
\multirow{2}{*}{\textbf{Tuned on}}
& \multirow{2}{*}{$(\lambda,\beta)$}
& \multicolumn{3}{c}{\textbf{WER (\%)}$\downarrow$} \\
\cmidrule(l){3-5}
& & \textbf{LRS3} & \textbf{LRS2} & \textbf{WildVSR} \\
\midrule

\rowcolor{gray!12}
\multicolumn{5}{@{}c@{}}{\textit{Implicit-length decoding (no length guidance)}} \\
-- & -- & 20.5 & 17.7 & 42.6 \\

\addlinespace[2pt]
\rowcolor{gray!12}
\multicolumn{5}{@{}c@{}}{\textit{Length-guided candidate decoding: reranking weights}} \\
LRS3 & $(0.9, 0.6)$ & 19.4 & 17.1 & 39.9 \\
LRS2 & $(1.0, 0.9)$ & 19.5 & 16.8 & 40.0 \\

\addlinespace[2pt]
\rowcolor{gray!12}
\multicolumn{5}{@{}c@{}}{\textit{WildVSR tuning (held-out analysis)}} \\
WildVSR$^\dagger$ & $(0.62, 0.32)$ & -- & -- & 40.0\,{\scriptsize$\pm$0.1} \\

\bottomrule
\end{tabular}
\end{adjustbox}
\caption{Cross-dataset robustness of reranking weights using USR~2.0 with $R=3$. Since WildVSR has no validation set, $^\dagger$ indicates that the weights are tuned on a held-out 10\% subset, and each tuned pair is evaluated on the full test set; mean $\pm$ std over five random splits.}
\vspace{-2pt}
\label{table:a1}
\end{table}

\subsection{Length Prediction Accuracy Across Datasets}
To further analyze the out-of-domain results on WildVSR, Table~\ref{table:6} evaluates the length predictor, trained only on LRS3, on both LRS3 and WildVSR. On LRS3, Acc@0 is around 44\%, Acc@1 around 88\%, and Acc@3 above 99\% for both encoders. On WildVSR, Acc@0 decreases to around 19\% and Acc@1 to around 50--52\%. However, the ground-truth length remains within $\pm3$ tokens of the prediction for about 85\% of samples and within $\pm5$ tokens for about 95--96\%. With the default search radius of $R=3$, the ground-truth length is therefore included in the candidate range for about 85\% of WildVSR samples, supporting length-guided candidate decoding under the out-of-domain setting.

\begin{table}[t]
\centering
\small
\setlength{\tabcolsep}{4pt}
\renewcommand{\arraystretch}{1.2}
\begin{adjustbox}{max width=\linewidth}
\begin{tabular}{@{}cccccc@{}}
\toprule
\multirow{2}{*}{\textbf{$R$}} &
\multirow{2}{*}{\textbf{\# Cand.}} &
\multicolumn{2}{c}{\textbf{WER (\%)}$\downarrow$} &
\multicolumn{2}{c}{\textbf{RTF}$\downarrow$} \\
\cmidrule(l){3-4} \cmidrule(l){5-6}
& &
\textbf{AV-HuBERT} &
\textbf{USR~2.0} &
\textbf{AV-HuBERT} &
\textbf{USR~2.0} \\
\midrule

\rowcolor{gray!12}
\multicolumn{6}{@{}c@{}}{\textit{Implicit-length decoding (no length guidance)}} \\
-- & 1 & 23.1 & 20.5 & 0.14 & 0.14 \\

\addlinespace[2pt]
\rowcolor{gray!12}
\multicolumn{6}{@{}c@{}}{\textit{Length-guided candidate decoding}} \\
1 & 3  & 22.6 & 20.3 & 0.31 & 0.29 \\
2 & 5  & 22.1 & 19.7 & 0.60 & 0.55 \\
\rowcolor{gray!8}
3 & 7  & 21.9 & 19.4 & 0.91 & 0.87 \\
5 & 11 & 21.9 & 19.5 & 1.59 & 1.51 \\

\bottomrule
\end{tabular}
\end{adjustbox}
\caption{Accuracy--efficiency trade-off with varying candidate-length search radius $R$ on LRS3. We use $R=3$ as the default setting.}
\vspace{-2pt}
\label{table:7}
\end{table}
\begin{table}[t]
\centering
\small
\setlength{\tabcolsep}{5pt}
\renewcommand{\arraystretch}{1.08}
\begin{tabularx}{\linewidth}{@{}lX@{}}
\toprule
\textbf{Step / Type} & \textbf{Generated transcript} \\
\midrule
\rowcolor{gray!12}
\multicolumn{2}{@{}l@{}}{\textit{Success case: different generation orders}} \\
\rowcolor{gray!6}
\multicolumn{2}{@{}l@{}}{\textit{Strict left-to-right order}} \\
Step 2  & \texttt{sage vision} \\
Step 6  & \texttt{sage vision is one of} \\
Step 11 & \texttt{sage vision is one of the greatest and most effective} \\
Step 16 & \texttt{sage vision is one of the greatest and most effective forms of political writing} \\

\rowcolor{gray!6}
\multicolumn{2}{@{}l@{}}{\textit{Full-parallel order}} \\
Step 1 & \texttt{\_ \_ \_ \_ \_ \_ \_ and \_ \_ \_ \_ \_ \_} \\
Step 2 & \texttt{\_ \_ is one of the greatest and most effective forms of political writing} \\
Step 3 & \texttt{\_ fiction is one of the greatest and most effective forms of political writing} \\
Step 4 & \texttt{science fiction is one of the greatest and most effective forms of political writing} \\

\midrule
\rowcolor{gray!12}
\multicolumn{2}{@{}l@{}}{\textit{Failure cases}} \\
Case 1
& GT: \texttt{harvest contaminated tobacco} \newline
  Pred: \texttt{harvested contaminated tobacco}
  \quad (\texttt{harvest} $\rightarrow$ \texttt{harvested}) \\
Case 2
& GT: \texttt{and remember we had experience with this} \newline
  Pred: \texttt{and i remembered we had experience with this}
  \quad (\texttt{remember} $\rightarrow$ \texttt{remembered}) \\
\bottomrule
\end{tabularx}
\caption{Qualitative comparison of token commitment orders with representative success and failure cases.}
\vspace{-4pt}
\label{table:8}
\end{table}

\subsection{Effect of Candidate-Length Search Radius}
\label{sec:radius}
Table~\ref{table:7} analyzes the accuracy--efficiency trade-off of length-guided candidate decoding as the candidate-length search radius $R$ varies. For this analysis, we decode the $K_{\mathrm{pred}} \pm 5$ candidate pool once and restrict the candidate set during reranking for each $R$; the reported RTF corresponds to decoding only the candidates within the respective radius. Increasing $R$ from 1 to 3 reduces WER from 20.3\% to 19.4\% with USR~2.0 and from 22.6\% to 21.9\% with AV-HuBERT, while increasing RTF. Increasing $R$ further to 5 does not improve WER and incurs additional decoding cost. We therefore use $R=3$ as the default setting.

\subsection{Qualitative Analysis}
Table~\ref{table:8} shows qualitative success and failure cases. In the success case, left-to-right decoding commits the erroneous prefix \texttt{sage vision}, whereas full-parallel decoding first reveals reliable later context and recovers \texttt{science fiction}. This shows that flexible-order decoding can provide useful semantic constraints for ambiguous positions. The failure cases show that errors can remain when ground-truth and predicted phrases are both fluent and visually plausible, such as \texttt{harvest} versus \texttt{harvested} or \texttt{remember} versus \texttt{remembered}.

\section{Conclusion}
We presented DLLM-VSR, a diffusion large language model-based framework that revisits token commitment order for visual speech recognition. By replacing left-to-right autoregressive decoding with flexible-order masked denoising, the framework allows confident tokens to serve as contextual anchors for resolving visually ambiguous positions. We further introduced two-stage masked-denoising training and length-guided candidate decoding to address fixed-canvas training and length uncertainty. Experiments on LRS3 and LRS2 demonstrate consistent improvements over autoregressive baselines, while evaluation on WildVSR shows that these improvements extend to an out-of-domain setting.

\section*{Limitations}
We find that controlling token generation order substantially benefits DLLM-based VSR, and oracle-length decoding reveals the upper bound achievable when the transcript length is known. While our length-guided candidate decoding narrows the gap to this oracle setting, a gap remains, indicating that length modeling can be further improved. Moreover, evaluating multiple length hypotheses increases inference time, so reducing the computational overhead of length-guided candidate decoding is an important direction for future work. Our evaluation is limited to English benchmarks spanning curated and in-the-wild settings; robustness to more heterogeneous visual conditions and potential performance disparities across demographic and speaker-specific factors have not been evaluated at the subgroup level and remain open questions.

\section*{Ethical Considerations}
VSR systems can improve accessibility for deaf and hard-of-hearing users and enable silent communication, but they may raise privacy concerns because spoken content could be inferred from silent videos without speaker consent. Such systems should therefore be used only with appropriate consent and safeguards. We use existing datasets and pretrained models only for VSR research. Given potential performance disparities across speakers and demographic groups, we do not recommend deployment in high-stakes settings without careful validation and human oversight.

\section*{Acknowledgments}
This work was supported in part by IITP grant funded by the Korea government (MSIT) (No. RS-2020-II200004, Development of Previsional Intelligence based on Long-Term Visual Memory Network), in part by the National Research Foundation of Korea (NRF) grant funded by the Korea government (MSIT) (RS-2022-NR070162), and in part by the KAIST Jang Young Sil Fellow Program.

% Bibliography entries for the entire Anthology, followed by custom entries
%\bibliography{anthology,custom}
% Custom bibliography entries only
\bibliography{custom}

@article{assael2016lipnet,
  title={Lipnet: End-to-end sentence-level lipreading},
  author={Assael, Yannis M and Shillingford, Brendan and Whiteson, Shimon and De Freitas, Nando},
  journal={arXiv preprint arXiv:1611.01599},
  year={2016}
}

@inproceedings{zhao2020hearing,
  title={Hearing lips: Improving lip reading by distilling speech recognizers},
  author={Zhao, Ya and Xu, Rui and Wang, Xinchao and Hou, Peng and Tang, Haihong and Song, Mingli},
  booktitle={Proceedings of the AAAI Conference on Artificial Intelligence},
  volume={34},
  number={04},
  pages={6917--6924},
  year={2020}
}

@article{yeo2024akvsr,
  title={Akvsr: Audio knowledge empowered visual speech recognition by compressing audio knowledge of a pretrained model},
  author={Yeo, Jeong Hun and Kim, Minsu and Choi, Jeongsoo and Kim, Dae Hoe and Ro, Yong Man},
  journal={IEEE Transactions on Multimedia},
  volume={26},
  pages={6462--6474},
  year={2024},
  publisher={IEEE}
}

@inproceedings{
shi2022learning,
title={Learning Audio-Visual Speech Representation by Masked Multimodal Cluster Prediction},
author={Bowen Shi and Wei-Ning Hsu and Kushal Lakhotia and Abdelrahman Mohamed},
booktitle={International Conference on Learning Representations},
year={2022},
url={https://openreview.net/forum?id=Z1Qlm11uOM}
}

@article{afouras2018lrs3,
  title={LRS3-TED: a large-scale dataset for visual speech recognition},
  author={Afouras, Triantafyllos and Chung, Joon Son and Zisserman, Andrew},
  journal={arXiv preprint arXiv:1809.00496},
  year={2018}
}

@INPROCEEDINGS{prajwal2022sub,
  author={Prajwal, K R and Afouras, Triantafyllos and Zisserman, Andrew},
  booktitle={2022 IEEE/CVF Conference on Computer Vision and Pattern Recognition (CVPR)}, 
  title={Sub-word Level Lip Reading With Visual Attention}, 
  year={2022},
  volume={},
  number={},
  pages={5152-5162},
  doi={10.1109/CVPR52688.2022.00510}}

@inproceedings{makino2019recurrent,
  title={Recurrent neural network transducer for audio-visual speech recognition},
  author={Makino, Takaki and Liao, Hank and Assael, Yannis and Shillingford, Brendan and Garcia, Basilio and Braga, Otavio and Siohan, Olivier},
  booktitle={2019 IEEE automatic speech recognition and understanding workshop (ASRU)},
  pages={905--912},
  year={2019},
  organization={IEEE}
}

@inproceedings{serdyuk2022transformer,
  title     = {{Transformer-Based Video Front-Ends for Audio-Visual Speech Recognition for Single and Multi-Person Video}},
  author    = {Dmitriy Serdyuk and Otavio Braga and Olivier Siohan},
  year      = {2022},
  booktitle = {{Interspeech 2022}},
  pages     = {2833--2837},
  doi       = {10.21437/Interspeech.2022-10920},
  issn      = {2958-1796},
}

@inproceedings{chang2024conformer,
  title={Conformer is all you need for visual speech recognition},
  author={Chang, Oscar and Liao, Hank and Serdyuk, Dmitriy and Shah, Ankit and Siohan, Olivier},
  booktitle={ICASSP 2024-2024 IEEE International Conference on Acoustics, Speech and Signal Processing (ICASSP)},
  pages={10136--10140},
  year={2024},
  organization={IEEE}
}

@ARTICLE{zhu2023vatlm,
  author={Zhu, Qiushi and Zhou, Long and Zhang, Ziqiang and Liu, Shujie and Jiao, Binxing and Zhang, Jie and Dai, Lirong and Jiang, Daxin and Li, Jinyu and Wei, Furu},
  journal={IEEE Transactions on Multimedia}, 
  title={VatLM: Visual-Audio-Text Pre-Training With Unified Masked Prediction for Speech Representation Learning}, 
  year={2024},
  volume={26},
  number={},
  pages={1055-1064},
  doi={10.1109/TMM.2023.3275873}}

@inproceedings{
haliassos2023jointly,
title={Jointly Learning Visual and Auditory Speech Representations from Raw Data},
author={Alexandros Haliassos and Pingchuan Ma and Rodrigo Mira and Stavros Petridis and Maja Pantic},
booktitle={The Eleventh International Conference on Learning Representations },
year={2023},
url={https://openreview.net/forum?id=BPwIgvf5iQ}
}

@inproceedings{stafylakis2017combining,
  title     = {{Combining Residual Networks with LSTMs for Lipreading}},
  author    = {Themos Stafylakis and Georgios Tzimiropoulos},
  year      = {2017},
  booktitle = {{Interspeech 2017}},
  pages     = {3652--3656},
  doi       = {10.21437/Interspeech.2017-85},
  issn      = {2958-1796},
}

@article{afouras2018deep,
  title={Deep audio-visual speech recognition},
  author={Afouras, Triantafyllos and Chung, Joon Son and Senior, Andrew and Vinyals, Oriol and Zisserman, Andrew},
  journal={IEEE transactions on pattern analysis and machine intelligence},
  volume={44},
  number={12},
  pages={8717--8727},
  year={2018},
  publisher={IEEE}
}

@INPROCEEDINGS{ren2021learning,
  author={Ren, Sucheng and Du, Yong and Lv, Jianming and Han, Guoqiang and He, Shengfeng},
  booktitle={2021 IEEE/CVF Conference on Computer Vision and Pattern Recognition (CVPR)}, 
  title={Learning from the Master: Distilling Cross-modal Advanced Knowledge for Lip Reading}, 
  year={2021},
  volume={},
  number={},
  pages={13320-13328},
  doi={10.1109/CVPR46437.2021.01312}}

@inproceedings{ma2023auto,
  title={Auto-AVSR: Audio-visual speech recognition with automatic labels},
  author={Ma, Pingchuan and Haliassos, Alexandros and Fernandez-Lopez, Adriana and Chen, Honglie and Petridis, Stavros and Pantic, Maja},
  booktitle={ICASSP 2023-2023 IEEE International Conference on Acoustics, Speech and Signal Processing (ICASSP)},
  pages={1--5},
  year={2023},
  organization={IEEE}
}

@inproceedings{
haliassos2026pay,
title={Pay Attention to {CTC}: Fast and Robust Pseudo-Labelling for Unified Speech Recognition},
author={Alexandros Haliassos and Rodrigo Mira and Stavros Petridis},
booktitle={The Fourteenth International Conference on Learning Representations},
year={2026},
url={https://openreview.net/forum?id=sSbEEHNEsL}
}

@inproceedings{yeo2024visual,
  title={Where visual speech meets language: VSP-LLM framework for efficient and context-aware visual speech processing},
  author={Yeo, Jeong Hun and Han, Seunghee and Kim, Minsu and Ro, Yong Man},
  booktitle={Findings of the Association for Computational Linguistics: EMNLP 2024},
  pages={11391--11406},
  year={2024}
}

@inproceedings{cappellazzo2025large,
  title={Large language models are strong audio-visual speech recognition learners},
  author={Cappellazzo, Umberto and Kim, Minsu and Chen, Honglie and Ma, Pingchuan and Petridis, Stavros and Falavigna, Daniele and Brutti, Alessio and Pantic, Maja},
  booktitle={ICASSP 2025-2025 IEEE International Conference on Acoustics, Speech and Signal Processing (ICASSP)},
  pages={1--5},
  year={2025},
  organization={IEEE}
}

@inproceedings{prajwal2024speech,
  title={Speech Recognition Models are Strong Lip-readers},
  author={Prajwal, KR and Afouras, Triantafyllos and Zisserman, Andrew},
  booktitle={Proc. Interspeech 2024},
  pages={2425--2429},
  year={2024}
}

@InProceedings{Yuan_2025_ICCV,
    author    = {Yuan, Zhaoxin and Yang, Shuang and Shan, Shiguang and Chen, Xilin},
    title     = {Not Only Vision: Evolve Visual Speech Recognition via Peripheral Information},
    booktitle = {Proceedings of the IEEE/CVF International Conference on Computer Vision (ICCV)},
    month     = {October},
    year      = {2025},
    pages     = {3091-3100}
}

@inproceedings{cappellazzo2026omni,
  title={Omni-avsr: Towards unified multimodal speech recognition with large language models},
  author={Cappellazzo, Umberto and Liu, Xubo and Ma, Pingchuan and Petridis, Stavros and Pantic, Maja},
  booktitle={ICASSP 2026-2026 IEEE International Conference on Acoustics, Speech and Signal Processing (ICASSP)},
  pages={17772--17776},
  year={2026},
  organization={IEEE}
}

@misc{grattafiori2024llama,
      title={The Llama 3 Herd of Models}, 
      author={Aaron Grattafiori and Abhimanyu Dubey and Abhinav Jauhri and Abhinav Pandey and Abhishek Kadian and Ahmad Al-Dahle and Aiesha Letman and Akhil Mathur and Alan Schelten and Alex Vaughan and Amy Yang and Angela Fan and Anirudh Goyal and Anthony Hartshorn and Aobo Yang and Archi Mitra and Archie Sravankumar and Artem Korenev and Arthur Hinsvark and Arun Rao and Aston Zhang and Aurelien Rodriguez and Austen Gregerson and Ava Spataru and Baptiste Roziere and Bethany Biron and Binh Tang and Bobbie Chern and Charlotte Caucheteux and Chaya Nayak and Chloe Bi and Chris Marra and Chris McConnell and Christian Keller and Christophe Touret and Chunyang Wu and Corinne Wong and Cristian Canton Ferrer and Cyrus Nikolaidis and Damien Allonsius and Daniel Song and Danielle Pintz and Danny Livshits and Danny Wyatt and David Esiobu and Dhruv Choudhary and Dhruv Mahajan and Diego Garcia-Olano and Diego Perino and Dieuwke Hupkes and Egor Lakomkin and Ehab AlBadawy and Elina Lobanova and Emily Dinan and Eric Michael Smith and Filip Radenovic and Francisco Guzmán and Frank Zhang and Gabriel Synnaeve and Gabrielle Lee and Georgia Lewis Anderson and Govind Thattai and Graeme Nail and Gregoire Mialon and Guan Pang and Guillem Cucurell and Hailey Nguyen and Hannah Korevaar and Hu Xu and Hugo Touvron and Iliyan Zarov and Imanol Arrieta Ibarra and Isabel Kloumann and Ishan Misra and Ivan Evtimov and Jack Zhang and Jade Copet and Jaewon Lee and Jan Geffert and Jana Vranes and Jason Park and Jay Mahadeokar and Jeet Shah and Jelmer van der Linde and Jennifer Billock and Jenny Hong and Jenya Lee and Jeremy Fu and Jianfeng Chi and Jianyu Huang and Jiawen Liu and Jie Wang and Jiecao Yu and Joanna Bitton and Joe Spisak and Jongsoo Park and Joseph Rocca and Joshua Johnstun and Joshua Saxe and Junteng Jia and Kalyan Vasuden Alwala and Karthik Prasad and Kartikeya Upasani and Kate Plawiak and Ke Li and Kenneth Heafield and Kevin Stone and Khalid El-Arini and Krithika Iyer and Kshitiz Malik and Kuenley Chiu and Kunal Bhalla and Kushal Lakhotia and Lauren Rantala-Yeary and Laurens van der Maaten and Lawrence Chen and Liang Tan and Liz Jenkins and Louis Martin and Lovish Madaan and Lubo Malo and Lukas Blecher and Lukas Landzaat and Luke de Oliveira and Madeline Muzzi and Mahesh Pasupuleti and Mannat Singh and Manohar Paluri and Marcin Kardas and Maria Tsimpoukelli and Mathew Oldham and Mathieu Rita and Maya Pavlova and Melanie Kambadur and Mike Lewis and Min Si and Mitesh Kumar Singh and Mona Hassan and Naman Goyal and Narjes Torabi and Nikolay Bashlykov and Nikolay Bogoychev and Niladri Chatterji and Ning Zhang and Olivier Duchenne and Onur Çelebi and Patrick Alrassy and Pengchuan Zhang and Pengwei Li and Petar Vasic and Peter Weng and Prajjwal Bhargava and Pratik Dubal and Praveen Krishnan and Punit Singh Koura and Puxin Xu and Qing He and Qingxiao Dong and Ragavan Srinivasan and Raj Ganapathy and Ramon Calderer and Ricardo Silveira Cabral and Robert Stojnic and Roberta Raileanu and Rohan Maheswari and Rohit Girdhar and Rohit Patel and Romain Sauvestre and Ronnie Polidoro and Roshan Sumbaly and Ross Taylor and Ruan Silva and Rui Hou and Rui Wang and Saghar Hosseini and Sahana Chennabasappa and Sanjay Singh and Sean Bell and Seohyun Sonia Kim and Sergey Edunov and Shaoliang Nie and Sharan Narang and Sharath Raparthy and Sheng Shen and Shengye Wan and Shruti Bhosale and Shun Zhang and Simon Vandenhende and Soumya Batra and Spencer Whitman and Sten Sootla and Stephane Collot and Suchin Gururangan and Sydney Borodinsky and Tamar Herman and Tara Fowler and Tarek Sheasha and Thomas Georgiou and Thomas Scialom and Tobias Speckbacher and Todor Mihaylov and Tong Xiao and Ujjwal Karn and Vedanuj Goswami and Vibhor Gupta and Vignesh Ramanathan and Viktor Kerkez and Vincent Gonguet and Virginie Do and Vish Vogeti and Vítor Albiero and Vladan Petrovic and Weiwei Chu and Wenhan Xiong and Wenyin Fu and Whitney Meers and Xavier Martinet and Xiaodong Wang and Xiaofang Wang and Xiaoqing Ellen Tan and Xide Xia and Xinfeng Xie and Xuchao Jia and Xuewei Wang and Yaelle Goldschlag and Yashesh Gaur and Yasmine Babaei and Yi Wen and Yiwen Song and Yuchen Zhang and Yue Li and Yuning Mao and Zacharie Delpierre Coudert and Zheng Yan and Zhengxing Chen and Zoe Papakipos and Aaditya Singh and Aayushi Srivastava and Abha Jain and Adam Kelsey and Adam Shajnfeld and Adithya Gangidi and Adolfo Victoria and Ahuva Goldstand and Ajay Menon and Ajay Sharma and Alex Boesenberg and Alexei Baevski and Allie Feinstein and Amanda Kallet and Amit Sangani and Amos Teo and Anam Yunus and Andrei Lupu and Andres Alvarado and Andrew Caples and Andrew Gu and Andrew Ho and Andrew Poulton and Andrew Ryan and Ankit Ramchandani and Annie Dong and Annie Franco and Anuj Goyal and Aparajita Saraf and Arkabandhu Chowdhury and Ashley Gabriel and Ashwin Bharambe and Assaf Eisenman and Azadeh Yazdan and Beau James and Ben Maurer and Benjamin Leonhardi and Bernie Huang and Beth Loyd and Beto De Paola and Bhargavi Paranjape and Bing Liu and Bo Wu and Boyu Ni and Braden Hancock and Bram Wasti and Brandon Spence and Brani Stojkovic and Brian Gamido and Britt Montalvo and Carl Parker and Carly Burton and Catalina Mejia and Ce Liu and Changhan Wang and Changkyu Kim and Chao Zhou and Chester Hu and Ching-Hsiang Chu and Chris Cai and Chris Tindal and Christoph Feichtenhofer and Cynthia Gao and Damon Civin and Dana Beaty and Daniel Kreymer and Daniel Li and David Adkins and David Xu and Davide Testuggine and Delia David and Devi Parikh and Diana Liskovich and Didem Foss and Dingkang Wang and Duc Le and Dustin Holland and Edward Dowling and Eissa Jamil and Elaine Montgomery and Eleonora Presani and Emily Hahn and Emily Wood and Eric-Tuan Le and Erik Brinkman and Esteban Arcaute and Evan Dunbar and Evan Smothers and Fei Sun and Felix Kreuk and Feng Tian and Filippos Kokkinos and Firat Ozgenel and Francesco Caggioni and Frank Kanayet and Frank Seide and Gabriela Medina Florez and Gabriella Schwarz and Gada Badeer and Georgia Swee and Gil Halpern and Grant Herman and Grigory Sizov and Guangyi and Zhang and Guna Lakshminarayanan and Hakan Inan and Hamid Shojanazeri and Han Zou and Hannah Wang and Hanwen Zha and Haroun Habeeb and Harrison Rudolph and Helen Suk and Henry Aspegren and Hunter Goldman and Hongyuan Zhan and Ibrahim Damlaj and Igor Molybog and Igor Tufanov and Ilias Leontiadis and Irina-Elena Veliche and Itai Gat and Jake Weissman and James Geboski and James Kohli and Janice Lam and Japhet Asher and Jean-Baptiste Gaya and Jeff Marcus and Jeff Tang and Jennifer Chan and Jenny Zhen and Jeremy Reizenstein and Jeremy Teboul and Jessica Zhong and Jian Jin and Jingyi Yang and Joe Cummings and Jon Carvill and Jon Shepard and Jonathan McPhie and Jonathan Torres and Josh Ginsburg and Junjie Wang and Kai Wu and Kam Hou U and Karan Saxena and Kartikay Khandelwal and Katayoun Zand and Kathy Matosich and Kaushik Veeraraghavan and Kelly Michelena and Keqian Li and Kiran Jagadeesh and Kun Huang and Kunal Chawla and Kyle Huang and Lailin Chen and Lakshya Garg and Lavender A and Leandro Silva and Lee Bell and Lei Zhang and Liangpeng Guo and Licheng Yu and Liron Moshkovich and Luca Wehrstedt and Madian Khabsa and Manav Avalani and Manish Bhatt and Martynas Mankus and Matan Hasson and Matthew Lennie and Matthias Reso and Maxim Groshev and Maxim Naumov and Maya Lathi and Meghan Keneally and Miao Liu and Michael L. Seltzer and Michal Valko and Michelle Restrepo and Mihir Patel and Mik Vyatskov and Mikayel Samvelyan and Mike Clark and Mike Macey and Mike Wang and Miquel Jubert Hermoso and Mo Metanat and Mohammad Rastegari and Munish Bansal and Nandhini Santhanam and Natascha Parks and Natasha White and Navyata Bawa and Nayan Singhal and Nick Egebo and Nicolas Usunier and Nikhil Mehta and Nikolay Pavlovich Laptev and Ning Dong and Norman Cheng and Oleg Chernoguz and Olivia Hart and Omkar Salpekar and Ozlem Kalinli and Parkin Kent and Parth Parekh and Paul Saab and Pavan Balaji and Pedro Rittner and Philip Bontrager and Pierre Roux and Piotr Dollar and Polina Zvyagina and Prashant Ratanchandani and Pritish Yuvraj and Qian Liang and Rachad Alao and Rachel Rodriguez and Rafi Ayub and Raghotham Murthy and Raghu Nayani and Rahul Mitra and Rangaprabhu Parthasarathy and Raymond Li and Rebekkah Hogan and Robin Battey and Rocky Wang and Russ Howes and Ruty Rinott and Sachin Mehta and Sachin Siby and Sai Jayesh Bondu and Samyak Datta and Sara Chugh and Sara Hunt and Sargun Dhillon and Sasha Sidorov and Satadru Pan and Saurabh Mahajan and Saurabh Verma and Seiji Yamamoto and Sharadh Ramaswamy and Shaun Lindsay and Shaun Lindsay and Sheng Feng and Shenghao Lin and Shengxin Cindy Zha and Shishir Patil and Shiva Shankar and Shuqiang Zhang and Shuqiang Zhang and Sinong Wang and Sneha Agarwal and Soji Sajuyigbe and Soumith Chintala and Stephanie Max and Stephen Chen and Steve Kehoe and Steve Satterfield and Sudarshan Govindaprasad and Sumit Gupta and Summer Deng and Sungmin Cho and Sunny Virk and Suraj Subramanian and Sy Choudhury and Sydney Goldman and Tal Remez and Tamar Glaser and Tamara Best and Thilo Koehler and Thomas Robinson and Tianhe Li and Tianjun Zhang and Tim Matthews and Timothy Chou and Tzook Shaked and Varun Vontimitta and Victoria Ajayi and Victoria Montanez and Vijai Mohan and Vinay Satish Kumar and Vishal Mangla and Vlad Ionescu and Vlad Poenaru and Vlad Tiberiu Mihailescu and Vladimir Ivanov and Wei Li and Wenchen Wang and Wenwen Jiang and Wes Bouaziz and Will Constable and Xiaocheng Tang and Xiaojian Wu and Xiaolan Wang and Xilun Wu and Xinbo Gao and Yaniv Kleinman and Yanjun Chen and Ye Hu and Ye Jia and Ye Qi and Yenda Li and Yilin Zhang and Ying Zhang and Yossi Adi and Youngjin Nam and Yu and Wang and Yu Zhao and Yuchen Hao and Yundi Qian and Yunlu Li and Yuzi He and Zach Rait and Zachary DeVito and Zef Rosnbrick and Zhaoduo Wen and Zhenyu Yang and Zhiwei Zhao and Zhiyu Ma},
      year={2024},
      eprint={2407.21783},
      archivePrefix={arXiv},
      primaryClass={cs.AI},
      url={https://arxiv.org/abs/2407.21783}, 
}

@misc{Yang2024Qwen25TR,
  title        = {Qwen2.5 Technical Report},
  author       = {An Yang and Baosong Yang and Beichen Zhang and Binyuan Hui and
                  Bo Zheng and Bowen Yu and Chengyuan Li and Dayiheng Liu and
                  Fei Huang and Haoran Wei and Huan Lin and Jian Yang and
                  Jianhong Tu and Jianwei Zhang and Jianxin Yang and Jiaxi Yang and
                  Jingren Zhou and Junyang Lin and Kai Dang and Keming Lu and
                  Keqin Bao and Kexin Yang and Le Yu and Mei Li and Mingfeng Xue and
                  Pei Zhang and Qin Zhu and Rui Men and Runji Lin and Tianhao Li and
                  Tianyi Tang and Tingyu Xia and Xingzhang Ren and Xuancheng Ren and
                  Yang Fan and Yang Su and Yichang Zhang and Yu Wan and Yuqiong Liu and
                  Zeyu Cui and Zhenru Zhang and Zihan Qiu},
  year         = {2025},
  eprint       = {2412.15115},
  archivePrefix= {arXiv},
  primaryClass = {cs.CL},
  url          = {https://arxiv.org/abs/2412.15115}
}

@inproceedings{bear2016decoding,
  title={Decoding visemes: Improving machine lip-reading},
  author={Bear, Helen L and Harvey, Richard},
  booktitle={2016 IEEE International Conference on Acoustics, Speech and Signal Processing (ICASSP)},
  pages={2009--2013},
  year={2016},
  organization={IEEE}
}

@inproceedings{cappelletta2012phoneme,
  title={Phoneme-to-viseme mapping for visual speech recognition},
  author={Cappelletta, Luca and Harte, Naomi},
  booktitle={International Conference on Pattern Recognition Applications and Methods},
  volume={2},
  pages={322--329},
  year={2012},
  organization={SciTePress}
}

@inproceedings{nie2025large,
 author = {Nie, Shen and Zhu, Fengqi and You, Zebin and Zhang, Xiaolu and Ou, Jingyang and Hu, Jun and Zhou, Jun and Lin, Yankai and Wen, Ji-Rong and LI, Chongxuan},
 booktitle = {Advances in Neural Information Processing Systems},
 doi = {10.52202/085713-1689},
 editor = {D. Belgrave and C. Zhang and H. Lin and R. Pascanu and P. Koniusz and M. Ghassemi and N. Chen},
 pages = {50608--50646},
 publisher = {Curran Associates, Inc.},
 title = {Large Language Diffusion Models},
 url = {https://proceedings.neurips.cc/paper_files/paper/2025/file/48b383b24230e0e6e649d9c98dae4d8c-Paper-Conference.pdf},
 volume = {38, Main Conference},
 year = {2025}
}

@article{ye2025dream,
  title={Dream 7b: Diffusion large language models},
  author={Ye, Jiacheng and Xie, Zhihui and Zheng, Lin and Gao, Jiahui and Wu, Zirui and Jiang, Xin and Li, Zhenguo and Kong, Lingpeng},
  journal={arXiv preprint arXiv:2508.15487},
  year={2025}
}

@article{yu2025dimple,
  title={Dimple: Discrete diffusion multimodal large language model with parallel decoding},
  author={Yu, Runpeng and Ma, Xinyin and Wang, Xinchao},
  journal={arXiv preprint arXiv:2505.16990},
  year={2025}
}

@inproceedings{
li2026beyond,
title={Beyond Fixed: Training-Free Variable-Length Denoising for Diffusion Large Language Models},
author={Jinsong Li and Xiaoyi Dong and Yuhang Zang and Yuhang Cao and Jiaqi Wang and Dahua Lin},
booktitle={The Fourteenth International Conference on Learning Representations},
year={2026},
url={https://openreview.net/forum?id=Ic2A2gCseC}
}

@misc{xie2025dream,
      title={Dream-Coder 7B: An Open Diffusion Language Model for Code}, 
      author={Zhihui Xie and Jiacheng Ye and Lin Zheng and Jiahui Gao and Jingwei Dong and Zirui Wu and Xueliang Zhao and Shansan Gong and Xin Jiang and Zhenguo Li and Lingpeng Kong},
      year={2025},
      eprint={2509.01142},
      archivePrefix={arXiv},
      primaryClass={cs.CL},
      url={https://arxiv.org/abs/2509.01142}, 
}

@inproceedings{yeo2024visual2,
  title={Visual speech recognition for languages with limited labeled data using automatic labels from whisper},
  author={Yeo, Jeong Hun and Kim, Minsu and Watanabe, Shinji and Ro, Yong Man},
  booktitle={ICASSP 2024-2024 IEEE International Conference on Acoustics, Speech and Signal Processing (ICASSP)},
  pages={10471--10475},
  year={2024},
  organization={IEEE}
}

@inproceedings{ma2021end,
  title={End-to-end audio-visual speech recognition with conformers},
  author={Ma, Pingchuan and Petridis, Stavros and Pantic, Maja},
  booktitle={ICASSP 2021-2021 IEEE International Conference on Acoustics, Speech and Signal Processing (ICASSP)},
  pages={7613--7617},
  year={2021},
  organization={IEEE}
}

@INPROCEEDINGS{kim2023lip,
  author={Kim, Minsu and Yeo, Jeong Hun and Choi, Jeongsoo and Ro, Yong Man},
  booktitle={2023 IEEE/CVF International Conference on Computer Vision (ICCV)}, 
  title={Lip Reading for Low-resource Languages by Learning and Combining General Speech Knowledge and Language-specific Knowledge}, 
  year={2023},
  volume={},
  number={},
  pages={15313-15325},
  doi={10.1109/ICCV51070.2023.01409}}

@inproceedings{li2025lavida,
 author = {Li, Shufan and Kallidromitis, Konstantinos and Bansal, Hritik and Gokul, Akash and Kato, Yusuke and Kozuka, Kazuki and Kuen, Jason and Lin, Zhe and Chang, Kai-Wei and Grover, Aditya},
 booktitle = {Advances in Neural Information Processing Systems},
 doi = {10.52202/085713-3511},
 editor = {D. Belgrave and C. Zhang and H. Lin and R. Pascanu and P. Koniusz and M. Ghassemi and N. Chen},
 pages = {105101--105134},
 publisher = {Curran Associates, Inc.},
 title = {LaViDa: A Large Diffusion Language Model for Multimodal Understanding},
 url = {https://proceedings.neurips.cc/paper_files/paper/2025/file/975affbe7b5f3b55fba62247b6877b1c-Paper-Conference.pdf},
 volume = {38, Main Conference},
 year = {2025}
}

@inproceedings{gong2025scaling,
 author = {Gong, Shansan and Agarwal, Shivam and Zhang, Yizhe and Ye, Jiacheng and Zheng, Lin and Li, Mukai and An, Chenxin and Zhao, Peilin and Bi, Wei and Han, Jiawei and Peng, Hao and Kong, Lingpeng},
 booktitle = {International Conference on Learning Representations},
 editor = {Y. Yue and A. Garg and N. Peng and F. Sha and R. Yu},
 pages = {5046--5073},
 title = {Scaling Diffusion Language Models via Adaptation from Autoregressive Models},
 url = {https://proceedings.iclr.cc/paper_files/paper/2025/file/0fa81c3f0d57f95b8776de3a248ef0ed-Paper-Conference.pdf},
 volume = {2025},
 year = {2025}
}

@InProceedings{you2025llada,
    author    = {You, Zebin and Nie, Shen and Zhang, Xiaolu and ZHOU, JUN and Lu, Zhiwu and Wen, Ji-Rong and Li, Chongxuan},
    title     = {LLaDA-V: Large Language Diffusion Models with Visual Instruction Tuning},
    booktitle = {Proceedings of the IEEE/CVF Conference on Computer Vision and Pattern Recognition (CVPR)},
    month     = {June},
    year      = {2026},
    pages     = {10093-10105}
}

@article{cheng2026improving,
  title={Improving Variable-Length Generation in Diffusion Language Models via Length Regularization},
  author={Cheng, Zicong and Jia, Ruixuan and Li, Jia and Yang, Guo-Wei and Guo, Meng-Hao and Hu, Shi-Min},
  journal={arXiv preprint arXiv:2602.07546},
  year={2026}
}

@article{yang2025diffusion,
  title={Diffusion llm with native variable generation lengths: Let [eos] lead the way},
  author={Yang, Yicun and Wang, Cong and Wang, Shaobo and Wen, Zichen and Qi, Biqing and Xu, Hanlin and Zhang, Linfeng},
  journal={arXiv preprint arXiv:2510.24605},
  year={2025}
}

@inproceedings{wu2026dreamon,
 author = {Wu, Zirui and Zheng, Lin and Xie, Zhihui and Ye, Jiacheng and Gao, Jiahui and Gong, Shansan and Feng, Yansong and Li, Zhenguo and BI, Wei and Zhou, Guorui and Kong, Lingpeng},
 booktitle = {International Conference on Learning Representations},
 editor = {C. Vondrick and B. Hariharan and C. Raffel and L. Pinto and D. Yang and A. Faust},
 pages = {76626--76643},
 title = {DreamOn: Diffusion Language Models For Code Infilling Beyond Fixed-size Canvas},
 url = {https://proceedings.iclr.cc/paper_files/paper/2026/file/7c020dd308d7413b5078686723b2f857-Paper-Conference.pdf},
 volume = {2026},
 year = {2026}
}

@inproceedings{arriola2025block,
  title={Block diffusion: Interpolating between autoregressive and diffusion language models},
  author={Arriola, Marianne and Gokaslan, Aaron and Chiu, Justin and Yang, Zhihan and Qi, Zhixuan and Han, Jiaqi and Sahoo, Subham and Kuleshov, Volodymyr},
  booktitle={International Conference on Learning Representations},
  volume={2025},
  pages={50726--50753},
  year={2025}
}

@inproceedings{jain2026hype,
  title={From hype to insight: Rethinking large language model integration in visual speech recognition},
  author={Jain, Rishabh and Harte, Naomi},
  booktitle={ICASSP 2026-2026 IEEE International Conference on Acoustics, Speech and Signal Processing (ICASSP)},
  pages={18717--18721},
  year={2026},
  organization={IEEE}
}

@inproceedings{djilali2024vsr,
  title={Do vsr models generalize beyond lrs3?},
  author={Djilali, Yasser Abdelaziz Dahou and Narayan, Sanath and LeBihan, Eustache and Boussaid, Haithem and Almazrouei, Ebtesam and Debbah, Merouane},
  booktitle={Proceedings of the IEEE/CVF Winter Conference on Applications of Computer Vision},
  pages={6635--6644},
  year={2024}
}

\clearpage
\appendix
\appendix

\begin{table*}[t!]
\centering
\scriptsize
\setlength{\tabcolsep}{3pt}
\renewcommand{\arraystretch}{1.05}
\begin{tabular}{@{}lccccccccccc@{}}
\toprule
\multicolumn{12}{@{}l@{}}{\textit{LRS3 AV-HuBERT validation set}} \\
\midrule
$\lambda \backslash \beta$
& 0.0 & 0.1 & 0.2 & 0.3 & 0.4 & 0.5 & 0.6 & 0.7 & 0.8 & 0.9 & 1.0 \\
\midrule
0.0 & 15.76 & 15.24 & 15.24 & 15.53 & 15.78 & 15.78 & 15.91 & 16.06 & 16.26 & 16.45 & 16.55 \\
0.1 & 13.28 & 13.30 & 13.34 & 13.72 & 13.94 & 14.33 & 14.63 & 14.80 & 15.03 & 15.09 & 15.18 \\
0.2 & 12.63 & 12.45 & 12.74 & 12.84 & 13.13 & 13.48 & 13.64 & 13.92 & 14.05 & 14.16 & 14.42 \\
0.3 & 12.38 & 12.16 & 12.22 & 12.37 & 12.65 & 12.71 & 12.91 & 13.21 & 13.32 & 13.54 & 13.64 \\
0.4 & 12.38 & 11.96 & 12.02 & 12.12 & 12.29 & 12.39 & 12.60 & 12.78 & 12.86 & 13.02 & 13.15 \\
0.5 & 12.38 & 11.92 & 11.93 & 11.99 & 12.02 & 12.15 & 12.24 & 12.39 & 12.67 & 12.77 & 12.80 \\
0.6 & 12.50 & 12.06 & 11.94 & 11.91 & 11.93 & 11.92 & 12.04 & 12.13 & 12.27 & 12.53 & 12.65 \\
0.7 & 12.49 & 12.02 & 11.84 & 11.80 & 11.84 & 11.83 & 11.91 & 11.90 & 11.96 & 12.13 & 12.35 \\
0.8 & 12.50 & 12.08 & 11.88 & 11.85 & 11.78 & 11.74 & 11.77 & 11.78 & 11.81 & 11.94 & 12.12 \\
0.9 & 12.56 & 12.17 & 11.89 & 11.87 & 11.79 & 11.76 & 11.77 & \textbf{11.73} & 11.77 & 11.82 & 11.95 \\
1.0 & 12.64 & 12.30 & 12.06 & 11.92 & 11.80 & 11.77 & 11.78 & 11.77 & 11.85 & 11.82 & 11.91 \\
\bottomrule
\end{tabular}
\caption{
Validation WER (\%) of length-guided reranking weights on LRS3 with AV-HuBERT. The best setting is $(\lambda,\beta)=(0.9,0.7)$, with a validation WER of 11.73\%.
}
\label{table:appendix_rerank_lrs3_avhubert}
\end{table*}

\begin{table*}[t!]
\centering
\scriptsize
\setlength{\tabcolsep}{3pt}
\renewcommand{\arraystretch}{1.05}
\begin{tabular}{@{}lccccccccccc@{}}
\toprule
\multicolumn{12}{@{}l@{}}{\textit{LRS3 USR~2.0 validation set}} \\
\midrule
$\lambda \backslash \beta$
& 0.0 & 0.1 & 0.2 & 0.3 & 0.4 & 0.5 & 0.6 & 0.7 & 0.8 & 0.9 & 1.0 \\
\midrule
0.0 & 19.57 & 19.24 & 19.63 & 19.87 & 19.96 & 20.38 & 20.85 & 21.01 & 21.16 & 21.31 & 21.55 \\
0.1 & 17.23 & 16.86 & 16.95 & 17.43 & 17.92 & 18.16 & 18.48 & 18.79 & 19.16 & 19.27 & 19.57 \\
0.2 & 16.92 & 16.51 & 16.60 & 16.63 & 17.01 & 17.24 & 17.73 & 17.86 & 18.18 & 18.37 & 18.55 \\
0.3 & 16.76 & 16.52 & 16.48 & 16.52 & 16.63 & 16.61 & 16.95 & 17.40 & 17.62 & 17.90 & 18.08 \\
0.4 & 16.85 & 16.34 & 16.27 & 16.29 & 16.41 & 16.44 & 16.61 & 16.73 & 17.16 & 17.40 & 17.57 \\
0.5 & 16.97 & 16.41 & 16.09 & 16.20 & 16.18 & 16.33 & 16.39 & 16.48 & 16.62 & 16.95 & 17.24 \\
0.6 & 17.06 & 16.47 & 16.24 & 16.17 & 16.16 & 16.19 & 16.27 & 16.34 & 16.45 & 16.55 & 16.71 \\
0.7 & 17.22 & 16.63 & 16.30 & 16.17 & 16.14 & 16.12 & 16.19 & 16.28 & 16.29 & 16.36 & 16.53 \\
0.8 & 17.31 & 16.70 & 16.42 & 16.15 & 16.11 & 16.15 & 16.18 & 16.22 & 16.29 & 16.28 & 16.40 \\
0.9 & 17.46 & 16.76 & 16.55 & 16.17 & 16.11 & 16.08 & \textbf{16.04} & 16.20 & 16.28 & 16.39 & 16.38 \\
1.0 & 17.43 & 16.94 & 16.55 & 16.20 & 16.11 & 16.11 & 16.05 & 16.09 & 16.32 & 16.32 & 16.37 \\
\bottomrule
\end{tabular}
\caption{
Validation WER (\%) of length-guided reranking weights on LRS3 with USR~2.0. The best setting is $(\lambda,\beta)=(0.9,0.6)$, with a validation WER of 16.04\%.
}
\label{table:appendix_rerank_lrs3_usr}
\end{table*}

\begin{table*}[t!]
\centering
\scriptsize
\setlength{\tabcolsep}{3pt}
\renewcommand{\arraystretch}{1.05}
\begin{tabular}{@{}lccccccccccc@{}}
\toprule
\multicolumn{12}{@{}l@{}}{\textit{LRS2 USR~2.0 validation set}} \\
\midrule
$\lambda \backslash \beta$
& 0.0 & 0.1 & 0.2 & 0.3 & 0.4 & 0.5 & 0.6 & 0.7 & 0.8 & 0.9 & 1.0 \\
\midrule
0.0 & 30.29 & 30.64 & 31.49 & 32.23 & 33.35 & 33.89 & 34.67 & 35.42 & 35.65 & 35.95 & 36.12 \\
0.1 & 24.80 & 24.78 & 25.65 & 26.51 & 26.89 & 27.97 & 28.68 & 29.24 & 30.42 & 30.98 & 31.63 \\
0.2 & 24.03 & 23.62 & 23.84 & 24.33 & 24.75 & 25.21 & 25.73 & 26.43 & 27.19 & 27.98 & 28.57 \\
0.3 & 23.75 & 23.35 & 23.19 & 23.46 & 23.74 & 23.85 & 24.40 & 24.90 & 25.17 & 25.64 & 26.28 \\
0.4 & 23.89 & 23.44 & 23.15 & 23.16 & 23.21 & 23.33 & 23.53 & 23.96 & 24.40 & 24.50 & 24.89 \\
0.5 & 23.98 & 23.42 & 23.29 & 23.16 & 23.20 & 23.09 & 23.12 & 23.46 & 23.66 & 23.98 & 24.13 \\
0.6 & 23.90 & 23.49 & 23.24 & 23.07 & 23.07 & 23.06 & 23.07 & 23.15 & 23.21 & 23.40 & 23.71 \\
0.7 & 23.90 & 23.61 & 23.35 & 23.24 & 23.07 & 22.96 & 23.01 & 23.05 & 23.10 & 23.09 & 23.20 \\
0.8 & 23.99 & 23.58 & 23.34 & 23.19 & 23.20 & 22.93 & 22.85 & 22.96 & 23.02 & 22.92 & 22.96 \\
0.9 & 24.08 & 23.66 & 23.47 & 23.24 & 23.18 & 23.09 & 23.00 & 22.97 & 22.87 & 22.92 & 22.87 \\
1.0 & 24.15 & 23.76 & 23.54 & 23.37 & 23.18 & 23.15 & 23.01 & 22.98 & 22.88 & \textbf{22.83} & 22.91 \\
\bottomrule
\end{tabular}
\caption{Validation WER (\%) of length-guided reranking weights on LRS2 with USR~2.0. The best setting is $(\lambda,\beta)=(1.0,0.9)$, with a validation WER of 22.83\%.}
\label{table:appendix_rerank_lrs2_usr}
\end{table*}

\section{Additional Experiments}
\label{appendix:additional}

\subsection{Cross-Dataset Robustness of Reranking Weights}
\label{appendix:rerank_crossdataset}

To examine the robustness of the reranking weights $(\lambda,\beta)$ across datasets, Table~\ref{table:a1} uses implicit-length decoding without length guidance as a reference. Applying the LRS3-tuned weights directly to LRS2 and WildVSR reduces WER from 17.7\% to 17.1\% and from 42.6\% to 39.9\%, respectively. Similarly, applying the LRS2-tuned weights to LRS3 and WildVSR reduces WER from 20.5\% to 19.5\% and from 42.6\% to 40.0\%, respectively. To further examine whether tuning directly on WildVSR provides additional benefit, we conduct a held-out analysis because WildVSR does not provide a validation set. We tune the weights on a randomly held-out 10\% subset, and evaluate each tuned pair on the full test set, obtaining 40.0\% WER with a standard deviation of 0.1 over five random splits. This result is comparable to the 39.9\% and 40.0\% WER obtained using the LRS3- and LRS2-tuned weights, indicating that the reranking weights remain effective without dataset-specific tuning.

\subsection{Behavior under Low-Confidence Conditions}
To examine how DLLM-VSR behaves when visual evidence does not support confident token commitment, we analyze decoding behavior under low-confidence conditions. Specifically, we consider cases where confidence remains low across the sequence. We first compute an utterance-level mean commit confidence by averaging the confidence of each transcript token at the step when it is committed. This confidence measure is negatively correlated with WER, with Pearson $r=-0.73$ on LRS3 and $-0.81$ on WildVSR, indicating that utterances with lower commit confidence tend to have higher recognition error. We further compare DLLM-VSR with an LLM-based VSR baseline~\cite{cappellazzo2025large} on the low-confidence subset, defined as utterances with mean commit confidence below 0.5. On LRS3, this subset contains 29 utterances (2.2\%), where DLLM-VSR obtains 90.1\% WER compared with 119.8\% for the baseline. On WildVSR, the subset contains 158 utterances (5.5\%), with WERs of 87.6\% and 161.0\%, respectively. In these subsets, 38\% of the baseline outputs exceed 100\% WER due to repetitive hallucinations, compared with 3--9\% for DLLM-VSR. The fixed-length canvas constrains the maximum output length, limiting such repetitive generation under low-confidence conditions. We also examine the prevalence of low-confidence cases. On LRS3, utterances for which more than 80\% of tokens have confidence below 0.5 account for only 0.5\% of the test set. Low-confidence cases are more frequent among short utterances, with a rate of 16.1\% for utterances of four words or fewer compared with 7.4\% for longer utterances.

\subsection{Reranking Weight Selection}
\label{appendix:rerank}

We analyze the selection of the balancing weights $\lambda$ and $\beta$ in the joint reranking score. Recall that $\lambda$ controls the contribution of the length-prediction score, while $\beta$ controls the iteration penalty. We select $(\lambda,\beta)$ by grid search over $[0,1]\times[0,1]$ with a step size of 0.1 on the validation set. Tables~\ref{table:appendix_rerank_lrs3_avhubert}--\ref{table:appendix_rerank_lrs2_usr} report the full validation grids for the three evaluated settings. The selected values are $(0.9,0.7)$ for AV-HuBERT/LRS3, $(0.9,0.6)$ for USR~2.0/LRS3, and $(1.0,0.9)$ for USR~2.0/LRS2, yielding validation WERs of 11.73\%, 16.04\%, and 22.83\%, respectively. Across the three settings, the selected values fall within $\lambda \in [0.9,1.0]$ and $\beta \in [0.6,0.9]$.

\section{Implementation Details}
\label{appendix:impl}

\noindent \textbf{Data statistics.}
In our processed datasets, the LRS3 training set (433 hours) consists of 164,916 utterances, of which 998 are held out as a validation set for tuning the reranking weights. The LRS3 test set contains 1,321 utterances. The LRS2 training set (223 hours) consists of 148,287 utterances, with 1,082 and 1,243 utterances used for validation and testing, respectively. WildVSR is used exclusively for evaluation and contains 2,854 test utterances.

\noindent \textbf{Pre-processing.}
We follow the standard VSR preprocessing protocol for LRS3 and LRS2. Video inputs are converted to greyscale mouth ROIs of size $88 \times 88$, using random cropping during training and center cropping during evaluation. Frames are sampled at 25 fps and normalized with mean 0.421 and standard deviation 0.165. During training, we apply temporal masking with a window of 10 frames and stride of 25 frames.

\noindent \textbf{Tokenization.}
Dream-7B and Qwen2.5-based experiments use the 152K Qwen2 tokenizer, with
\texttt{<|im\_end|>} as EOS and \texttt{<|image\_pad|>} as PAD. Inputs are
formatted with the standard Hugging Face \texttt{apply\_chat\_template} using
one user turn with the instruction: \textit{``You are given a silent video of
a speaker. Infer the spoken content from the lip movements and generate the
corresponding transcription.''}

\noindent \textbf{Architecture.}
The frozen visual encoder is either USR~2.0 Huge (32 layers, hidden size 1280) or AV-HuBERT Large (24 layers, hidden size 1024). Visual features are passed through a 1D convolutional length adapter with kernel size 2 and stride 2, followed by a two-layer projector consisting of LayerNorm--Linear--GELU--Linear. The projector uses a hidden dimension of $\lfloor(d_{\mathrm{enc}}+3584)/2\rfloor$ and maps the visual features to the 3584-dimensional DLLM hidden space. The base parameters of Dream-7B are frozen, while LoRA is applied to all linear layers with rank $r=16$, $\alpha=32$, and dropout 0.05. The total number of trainable parameters is 61.1M with USR~2.0, comprising 40.4M LoRA, 17.4M projector, and 3.3M length adapter parameters, and 53.1M with AV-HuBERT.

\begin{table}[t]
\centering
\small
\setlength{\tabcolsep}{4pt}
\renewcommand{\arraystretch}{1.08}
\begin{tabularx}{\linewidth}{@{}lX@{}}
\toprule
\textbf{Component} & \textbf{Setting} \\
\midrule
Visual encoder & USR~2.0 Huge / AV-HuBERT, frozen \\
Length adapter & Conv1d, kernel size 2, stride 2 \\
Projector & 2-layer MLP to 3584-dim LLM space \\
LLM decoder & Dream-7B, frozen \\
LoRA & $r=16$, $\alpha=32$, dropout 0.05, all linear layers \\
Canvas length & $T=32$ ($T=96$ for WildVSR) \\
Block size & 32 \\
Commit threshold & 0.9 \\
Candidate pool & $K_{\text{pred}} \pm 5$ (selection radius $R=3$) \\
\bottomrule
\end{tabularx}
\caption{Main implementation settings for DLLM-VSR.}
\label{table:appendix_impl}
\end{table}
\begin{table}[t]
\centering
\small
\setlength{\tabcolsep}{4pt}
\renewcommand{\arraystretch}{1.2}
\begin{adjustbox}{max width=\linewidth}
\begin{tabular}{@{}lccc@{}}
\toprule
\textbf{Component} & \textbf{Peak VRAM / GPU} & \textbf{Steps} & \textbf{Wall-clock} \\
\midrule
Stage-1 training   & 22.6 GB & 42,000 & 24.2 h \\
Stage-2 training   & 22.5 GB & 4,000  & 2.4 h  \\
Length predictor   & 10.5 GB & 18,000 & 4.2 h  \\
\midrule
Inference ($R=3$)  & 20.2 GB & --     & RTF 0.87 \\
\bottomrule
\end{tabular}
\end{adjustbox}
\caption{Measured computational cost of the full USR~2.0 pipeline. Training wall-clock times are on 8$\times$ RTX~3090 (24GB) with DeepSpeed ZeRO-2 and bf16 mixed precision; inference runs on a single RTX~3090. The complete pipeline takes approximately 31 hours (about 245 GPU-hours).}
\vspace{-2pt}
\label{tab:compute}
\end{table}

\noindent \textbf{Training.}
Both stages use AdamW with $\beta_1=0.9$, $\beta_2=0.999$, weight decay 0.01, and gradient clipping with maximum norm 1.0. Training uses bf16 mixed precision, DeepSpeed ZeRO-2 on 8 RTX 3090 GPUs, and dynamic batching with a maximum batch size of 32. Stage~1 is trained for 42K steps with a learning rate of $1\mathrm{e}{-4}$, and Stage~2 is fine-tuned from Stage~1 for 4K steps with a learning rate of $5\mathrm{e}{-5}$. We use a cosine scheduler with a minimum learning-rate scale of 0.1. Due to computational cost, each configuration is trained once, and all reported WER results are from a single run unless otherwise stated.

\noindent \textbf{Length predictor.}
The length predictor uses the same frozen visual encoder as the main model. It consists of a two-layer Transformer encoder with hidden dimension 384, 6 attention heads, feed-forward dimension 1536, and dropout 0.1. A learnable \texttt{[LEN]} token is prepended to the visual feature sequence, and its output is classified over candidate transcript lengths. The predictor has approximately 4M trainable parameters and is trained independently with cross-entropy loss on the target transcript length for 18K/5K steps with USR~2.0/AV-HuBERT on LRS3 and 20K steps on LRS2.

\noindent \textbf{Decoding and reranking.}
At inference, we use a fixed canvas length of $T=32$ ($T=96$ for WildVSR)
and a block size of 32. We decode the candidate pool within
$K_{\text{pred}} \pm 5$, resulting in up to 11 candidates; all candidates
are batched and decoded in parallel. During denoising, positions whose
confidence exceeds 0.9 are committed; if no position exceeds the threshold,
the most confident position is committed. The final transcript is selected
among candidates within radius $R=3$ of $K_{\text{pred}}$ using the joint
reranking score in Eq.~\ref{eq:rerank}; the search-radius analysis in
Sec.~\ref{sec:radius} varies only this selection radius. Reranking weights
$(\lambda,\beta)$ are selected by grid search over $[0,1]\times[0,1]$ with a
step size of 0.1 on the validation set. The selected values are $(0.9,0.7)$
for AV-HuBERT/LRS3, $(0.9,0.6)$ for USR~2.0/LRS3, and $(1.0,0.9)$ for
USR~2.0/LRS2.

\end{document}